\documentclass[lettersize,journal]{IEEEtran}
\usepackage{amsmath,amsfonts}
\usepackage{algorithmic}
\usepackage{algorithm}
\usepackage{array}
\usepackage[caption=false,font=normalsize,labelfont=sf,textfont=sf]{subfig}
\usepackage{textcomp}
\usepackage{stfloats}
\usepackage{url}
\usepackage{verbatim}
\usepackage{graphicx}
\usepackage{hyperref}
\usepackage{cite}
\usepackage{cleveref}
 \Crefname{figure}{Fig.}{Figs.}
\begin{document}

\title{3D-RCNet: Learning from Transformer to Build a 3D Relational ConvNet for Hyperspectral Image Classification}

\author{Haizhao Jing,
        Liuwei Wan, Xizhe Xue,
 Haokui Zhang*, 
 Ying Li}

% ~\IEEEmembership{Staff,~IEEE,}
        % <-this % stops a space
% \thanks{Haizhao Jing, Xizhe Xue, Haokui Zhang, Ying Li and Peng Wang are with the Northwestern Polytechnical University, Xi'an 710129, China (haizhao\_jing@mail.nwpu.edu.cn; xuexizhe@mail.nwpu.edu.cn; hkzhang@nwpu.edu.cn;lybyp@nwpu.edu.cn; peng.wang@nwpu.edu.cn).}% <-this % stops a space
% \thanks{Liuwei Wan is with the Yan'an University, Yan'an 716000, China (wanliuwei@yau.edu.cn).}% <-this % stops a space
% \thanks{Amirkhan Temirbayev is with the Al-Farabi Kazakh National University, Almaty, 050040, Kazakhstan (amirkhan@kaznu.kz).}

% \markboth{IEEE JOURNAL OF SELECTED TOPICS IN APPLIED EARTH OBSERVATIONS AND REMOTE SENSING,~Vol.~xx, 2024}%


\maketitle

\begin{abstract}
Recently, the Vision Transformer (ViT) model has replaced the classical Convolutional Neural Network (ConvNet) in various computer vision tasks due to its superior performance. Even in hyperspectral image (HSI) classification field, ViT-based methods also show promising potential. Nevertheless, ViT encounters notable difficulties in processing HSI data. Its self-attention mechanism, which exhibits quadratic complexity, escalates computational costs. Additionally, ViT's substantial demand for training samples does not align with the practical constraints posed by the expensive labeling of HSI data. To overcome these challenges, we propose a 3D relational ConvNet named 3D-RCNet, which inherits both strengths of ConvNet and ViT, resulting in high performance in HSI classification. We embed the self-attention mechanism of Transformer into the convolutional operation of ConvNet to design 3D relational convolutional operation and use it to build the final 3D-RCNet. The proposed 3D-RCNet maintains the high computational efficiency of ConvNet while enjoying the flexibility of ViT. Additionally, the proposed 3D relational convolutional operation is a plug-and-play operation, which can be inserted into previous ConvNet-based HSI classification methods seamlessly. Empirical evaluations on three representative benchmark HSI datasets show that the proposed model outperforms previous ConvNet-based and ViT-based HSI approaches. Github repository: \href{https://github.com/wanggynpuer/3D-RCNet}{https://github.com/wanggynpuer/3D-RCNet}.
\end{abstract}


\section{Introduction}
\IEEEPARstart{H}{yperspectral} image (HSI) consists of images captured across hundreds of spectral bands of a target area using sensors or imaging spectrometers, constituting three-dimensional data rich in spectral and complex spatial information. Due to the extensive information contained within HSI, its classification is widely applied across various domains such as geology and minerals \cite{2012Multi}\cite{2021HSIg}\cite{Guha2020}, atmospheric sciences \cite{Calin2021}, and agriculture \cite{Dale2013}\cite{Lu_Dao_Liu_He_Shang_2020}\cite{wang2021review}. The objective of HSI classification tasks is to accurately identify the classes of individual pixels in the HSI. Each spatial pixel in an HSI includes a complex array of high-dimensional spectral features, which while aiding in detailed classification, also increases the challenge of feature extraction. Thus, enhancing the robustness of feature extraction in HSI remains a primary research direction in the field of HSI classification.

With the rapid development of deep learning in the field of computer vision, an increasing number of research teams have designed deep learning-based feature extractors specifically tailored to the characteristics of HSI data. For instance, Qin et al. utilized graph convolutional networks (GCNs) to address the non-Euclidean structures in HSI data \cite{Qin2019Spectral}, while Mou et al. employed recurrent neural networks (RNNs) to effectively manage the long-range dependencies of spatial features \cite{2017Deep}. These deep learning-based feature extractors have shown promising results in HSI classification tasks. Notably, three-dimensional convolutional neural networks (3D ConvNet), with their inherently three-dimensional structure, are particularly adept at extracting features from HSI data. Chen et al. proposed a 3D ConvNet model to extract spectral-spatial features from HSI \cite{Chenyushi2016}, Li et al. developed a 3D ConvNet framework for HSI classification \cite{Liy2017}, and Zhang et al. introduced an end-to-end lightweight 3D ConvNet for HSI classification with limited samples \cite{Zhanghk2019}. These 3D ConvNet-based HSI feature extraction models can better preserve the intrinsic relationship between spatial and spectral features in the data when processing HSI data, and they all show good results in downstream task classification.

In 2020, Dosovitskiy et al. introduced Vision Transformer (ViT), marking the inception of image classification models based on the self-attention mechanism \cite{2021An}. Based on Transformer's ability to effectively handle long distance dependencies, ViT exhibited superior classification accuracy on the ImageNet-1k dataset compared to traditional ConvNet models. Inspired by this, researchers explored the application of ViT to HSI classification. Hong et al. were pioneers in applying a pure Transformer architecture to HSI classification, developing a novel model named SpectralFormer, which enhances group spectral embedding by assimilating local spectral sequence information from adjacent bands and facilitates the transfer of salient information throughout the extraction process via a multi-layer Transformer structure \cite{2021SpectralFormer}. Additionally, Sun et al. introduced a Transformer model that integrates spectral and spatial features, effectively extracting these features from HSI through tokenization \cite{SunLe}. Xin et al. proposed the Spa-Spe-TR model, employing a dual-branch pure Transformer framework to independently extract and subsequently fuse spectral and spatial information, thereby substantiating the efficacy of Transformers in HSI classification \cite{HeXin}.

In these explorations, to effectively harness ViT for global feature extraction from HSI data, we consider both the three-dimensional nature of HSI and the high computational complexity of Transformers. Typically, a hybrid structure combining 3D ConvNet and ViT is employed to enhance feature extraction performance. This design merges the capability of 3D ConvNet to efficiently process three-dimensional data with ViT's prowess in capturing global feature information. The result is a balanced hybrid structure that captures both local and global details while effectively reducing computational complexity. For instance, Sun et al. proposed a method for HSI classification that uses shallow layers of 3D ConvNet and 2D ConvNet to extract low-level semantic features, followed by deep Transformer modules for high-level semantic extraction \cite{Sun2024}. Similarly, Ghaderizadeh et al. introduced a multiscale dual-branch residual spectral-spatial network that utilizes convolutional kernels of varying sizes at shallow levels and Transformer modules at deeper levels to enhance semantic feature extraction \cite{Saeed2022}. These hybrid models, leveraging 3D ConvNet at the initial stages and ViT at more advanced stages, demonstrate superior performance in HSI classification tasks compared to traditional single-structure approaches. However, these models still primarily represent a direct concatenation of 3D ConvNet and ViT modules, which may retain some limitations of singular structure approaches.

In this work, we extend the design approach of combining 3D ConvNet and ViT structures by deeply understanding the structural advantages of both and proposing a 3D relational ConvNet named 3D-RCNet. We incorporate the Transformer’s self-attention mechanism into the convolutional operations of ConvNet, designing a 3D relational convolutional operation that forms the foundation of the final 3D-RCNet. The 3D relational convolutional operation extracts local spatial-spectral features through convolutional windows and applies attention operations within these windows to generate new features. Consequently, the constructed 3D-RCNet maintains the high computational efficiency of ConvNet while enjoying the flexibility of ViT for feature extraction. Moreover, the proposed 3D relational convolutional operation is a plug-and-play mechanism that can be seamlessly integrated into previous ConvNet-based HSI classification methods. Experimental results on three publicly available representative HSI datasets demonstrate that our proposed 3D-RCNet exhibits excellent performance in HSI classification tasks, outperforming previously proposed hybrid models.
The main contributions of this work are as follows:
\begin{enumerate}
    \item In this paper, taking the characteristic of HSI into consideration along with self-attention, we tailored a block named 3D Relational Convolutional Block(3D-RCBlock) where the self-attention mechanism is embedded into the convolutional operation, resulting in a new HSI feature extraction operation that inherits both strengths of ConvNet and ViT.
    \item Based on the proposed 3D-RCBlock, we have built a hybrid network framework in which the 3D-RCBlock is seamlessly integrated with a classical 3D ConvNet.
    \item We conducted exhaustive ablation experiments, analyzed each module of the structure in detail, and provided comprehensive guiding conclusions for the construction of the structure.
\end{enumerate}
\section{Related work}
\subsection{Hyperspectral Image Classification of 3D ConvNet}
With the widespread application of deep learning, HSI classification tasks have entered a phase of rapid development, wherein convolutional neural networks play a crucial role. This primarily includes methods using 1D ConvNet, 2D ConvNet, and 3D ConvNet. HSI classification methods based on 1D ConvNet and 2D ConvNet typically focus on either spectral or spatial information \cite{Mei2016}\cite{Yang2016}\cite{ 2017Spectral}\cite{Roy2020}\cite{Fang2019}, failing to effectively extract the rich spectral and complex spatial information contained in HSI data. Therefore, current research predominantly focuses on 3D ConvNet, which demonstrate better extraction performance for three-dimensional structured data. The convolutional kernels of 3D ConvNet slide along both spatial and spectral dimensions of HSI data, thus processing the spatial location and spectral band of each pixel during feature extraction. Liu et al.'s team validated the effectiveness of 3D ConvNet in feature extraction from HSI data through comparisons with 1D ConvNet and 2D ConvNet combined with principal component analysis \cite{Liu2018}. Ladi et al. demonstrated that the structure of 3D ConvNet exhibits superior performance in processing the three-dimensional data of HSI compared to 1D ConvNet and 2D ConvNet \cite{Ladi2022}. Chen et al. proposed a model combining hyperspectral target detection and 3D ConvNet, achieving effective HSI classification \cite{3dhsi2024}. Jiang et al. proposed a 3-D separable ResNet, which demonstrated excellent performance in HSI classification with small training samples\cite{Jiang_Li_Zhang_2019}.Due to its three-dimensional structure, 3D ConvNet can avoid the information loss caused by traditional convolutional downsampling methods and effectively extract both spectral and spatial features, making it widely used in HSI classification. Inspired by these, our proposed 3D-RCNet continues the module design similar to the 3D ConvNet three-dimensional structure at the shallow level, which is used to extract spectral and spatial features efficiently. However, 3D ConvNet use static convolutional kernels when processing different datasets, which limits their adaptability and flexibility in capturing features of diverse data distributions.

\subsection{Hyperspectral Image Classification of Vision Transformer}\label{sec:Title}
Dosovitskiy et al. first applied the Transformer model to computer vision tasks, proposing the Vision Transformer (ViT), which divides images into fixed-size patches, then flattens these patches into a one-dimensional sequence to be processed by the Transformer model\cite{2021An}. The ViT architecture is highly flexible, capable of accommodating images of different sizes and varying numbers of categories, and it achieves better performance than traditional ConvNet when trained on large-scale datasets.

Recently, with the deepening research of ViT in the field of machine vision, especially in image classification tasks, using ViT to extract HSI data features for classification has become a major research direction. Hong et al. proposed using grouped spectral embeddings instead of all spectral bands for feature extraction to model local detailed spectral differences\cite{2021SpectralFormer}. This work added cross-layer skip connections in multiple encoders of the Transformer to reduce information loss during layer-by-layer propagation. Sun et al. proposed a model combining spectral-spatial feature tokenization with the Transformer, improving the accuracy of HSI classification by segmenting and encoding spatial features of the images\cite{SunLe}. Jiang et al. proposed a Graph Generative Structure-Aware Transformer architecture that integrates spatial-spectral features and graph structure information in HSI classification to enhance classification performance\cite{Jiang2023}. In HSI classification tasks, ViT demonstrates excellent feature extraction performance for HSI data with different scale three-dimensional structures by leveraging self-attention mechanisms and global feature capturing capabilities. However, when training on HSI data, the ViT has a high demand for training samples, and the self-attention mechanism entails significant computational complexity. Additionally, due to the excessive flexibility of the self-attention mechanism, the training process may become unstable. 

\subsection{Hybrid Structures for Hyperspectral Image Classification}\label{sec:Title}
Both 3D ConvNet and ViT exhibit excellent performance in HSI classification tasks. However, due to the three-dimensional structural characteristics of HSI data and its complex spectral and spatial information, single 3D ConvNet and ViT models have certain application limitations. 3D ConvNet are better at handling three-dimensional structural data but cannot effectively manage long-range dependencies, making them more suitable for capturing low-level features of HSI data at shallow layers. ViT, with its attention mechanism, can handle long-range dependencies well, but its computational complexity is exponentially influenced by the scale of image data, making it more suitable for capturing spectral and spatial global information of HSI data at deeper layers. In their LeViT model, Graham et al. combined Transformer structures with ConvNet, optimizing the balance between inference speed and accuracy, significantly outperforming previous convolutional and ViT models\cite{Graham2021}. Fu et al. proposed IncepFormer, using deep convolutional and pooling for spatial reduction followed by attention operations, and employed a pyramid structure to capture rich features of objects of different sizes\cite{Fu2022}. In MobileViT, the deeper layers of MobileNetv2 were replaced with their proposed MobileViT blocks\cite{Mehta2021}. In ParCNet, Zhang et al. introduced Position-Aware Circular Convolutions (ParC) to capture global features and designed a hybrid structure by combining the proposed ParC operations with traditional convolutions\cite{ZhangH2022}.

Recently, researchers have introduced hybrid structures into HSI classification. Ghaderizadeh et al. proposed a Multi-Scale Dual-Branch Residual Spectral-Spatial Network (MDBRSSN), which uses 3D convolutions with different kernels in parallel within two branches to extract spatial-spectral features, subsequently inputting these features into an attention mechanism for feature extraction\cite{Ghaderizadeh2022}. This work fused the features extracted by the two branches, followed by 2D ConvNet and GAP for feature compression, and achieved good classification results after passing through a fully connected layer. The Spectral-Spatial Feature Tokenization Transformer (SSFTT) employs 3D ConvNet to extract shallow spectral features and 2D ConvNet to extract shallow spatial features, which are then Gaussian-weighted and tokenized before being input into the Transformer module for feature learning\cite{SunLe}, yielding effective results in HSI classification tasks. The T-SST-based classification framework (T-SST-L) uses ConvNet to extract spatial features and densely connected Transformer to capture sequential spectral relationships\cite{He2021}. These studies combine ConvNet and Transformer modules from a model structure perspective, effectively mixing ConvNet and Transformer modules to improve model accuracy.  

While previous hybrid models combining 3D ConvNet and ViT have shown promise in HSI classification, they typically rely on straightforward concatenation of these technologies, which can be limiting. Building on this foundational hybrid approach, our research takes a significant leap forward by more deeply integrating the convolutional capabilities of 3D ConvNet with the self-attention mechanisms of Transformers. This integration is realized through our novel 3D-RCNet model, which utilizes the specially designed 3D-RCBlock to seamlessly blend the strengths of both architectures, resulting in enhanced classification performance and operational efficiency.

\begin{figure*}[h!]
	\centering
	\includegraphics[width=\textwidth]{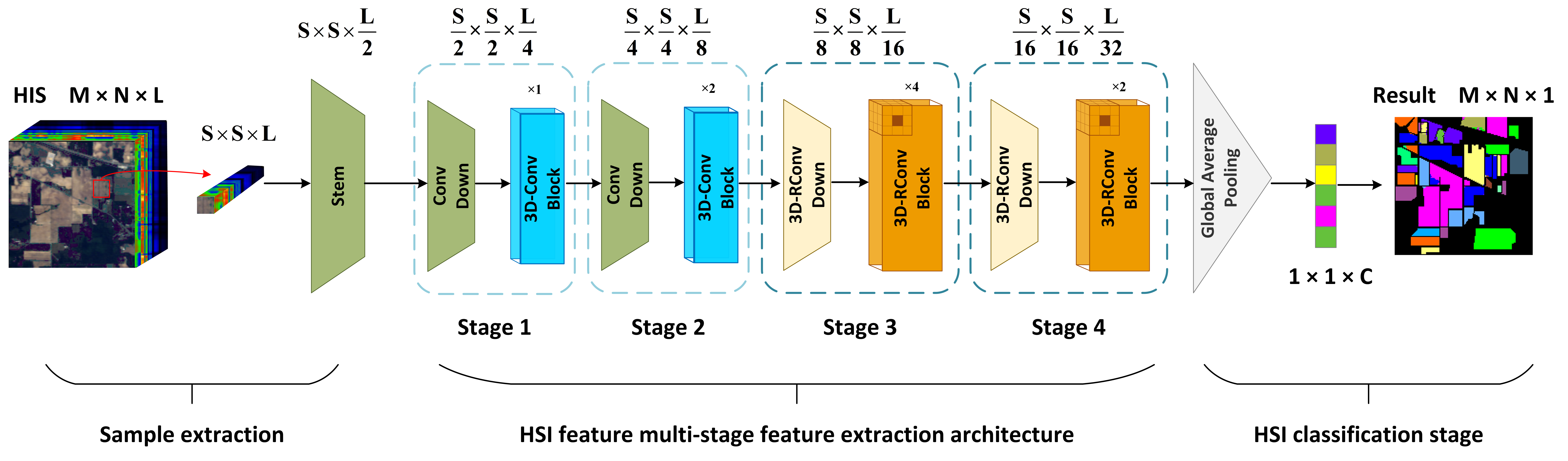}
	\caption{
		The 3D-RCNet framework proposed by us, and the framework uses four stages of blocks for feature extraction at different depths on HSI data.
	}
	\label{fig:overflow}
\end{figure*}
\section{Methodology}
\label{sec:methodology}
\subsection{3D-RCNet for HSI Classification}
\label{sec:3D-RCNet for HSI Classification}

\textbf{Network structure:} \Cref{fig:overflow} shows the overall framework of our proposed 3D-RCNet, which is designed for HSI classification and seamlessly integrates the 3D-RCBlock, equipped with a 3D relational convolutional operation. The framework is divided into three stages: the first stage is the sample extraction for HSI data; the second stage involves multi-layer feature extraction based on the 3D-RCNet framework; the third stage performs HSI classification using Global Average Pooling.

\textbf{Sample extraction:} In this stage, a data cube block with spatial dimensions S and spectral bands L is extracted as a sample, where the block size is defined as S × S × L. The block, centered around a pixel within its domain window in the HSI data, assigns the label of this central pixel to the entire data cube block. S is set as an odd number to ensure the central pixel is properly positioned for classification. Subsequently, this extracted data cube is input into the Stem layer, where the number of spectral bands is compressed and initial spectral information is extracted. 

\textbf{HSI feature multi-stage feature extraction architecture:} This multi-stage feature extraction architecture seamlessly incorporating 3D-RCBlock, utilizing a hybrid design of 3D ConvNet and 3D-RCBlock. In Stage 1, convolutional downsampling is applied to the input data to increase channel quantity while simultaneously reducing the resolution of the HSI to minimize feature information loss. Basic semantic-free features are extracted using a 3D-ConvBlock. Stage 2 continues with a similar setup, employing two 3D-ConvBlock to extract more complex and abstract features. In Stage 3 and Stage 4, 3D-RCBlock take over, initially reducing resolution and increasing channel count through 3D relational convolutional downsampling, and subsequently extracting more abstract and semantic features. This downsampling strategy across the four stages effectively reduces both spectral and spatial resolution at rates of [2,4,8,16], facilitating a progressive enhancement in feature representation. Experiments in \Cref{sec:Comparison Experiments} have shown that the framework, which utilizes 3D-ConvBlock in shallow layers and 3D-RCBlock in deeper layers, exhibits good performance. Moreover, ablation studies in \Cref{sec:Ablation Study} demonstrate that 3D-RCBlock can more effectively extract high-level semantic features in deeper layers.

\textbf{HSI classification stage:} By employing global average pooling, the entire feature map is compressed into a fixed-length vector. This global feature vector captures the comprehensive information of the image, which not only enables more efficient classification by the model but also enhances its ability to generalize to positional changes and unseen data within the input image.
\begin{figure*}[htbp]
	\centering
	\includegraphics[width=0.75\textwidth]{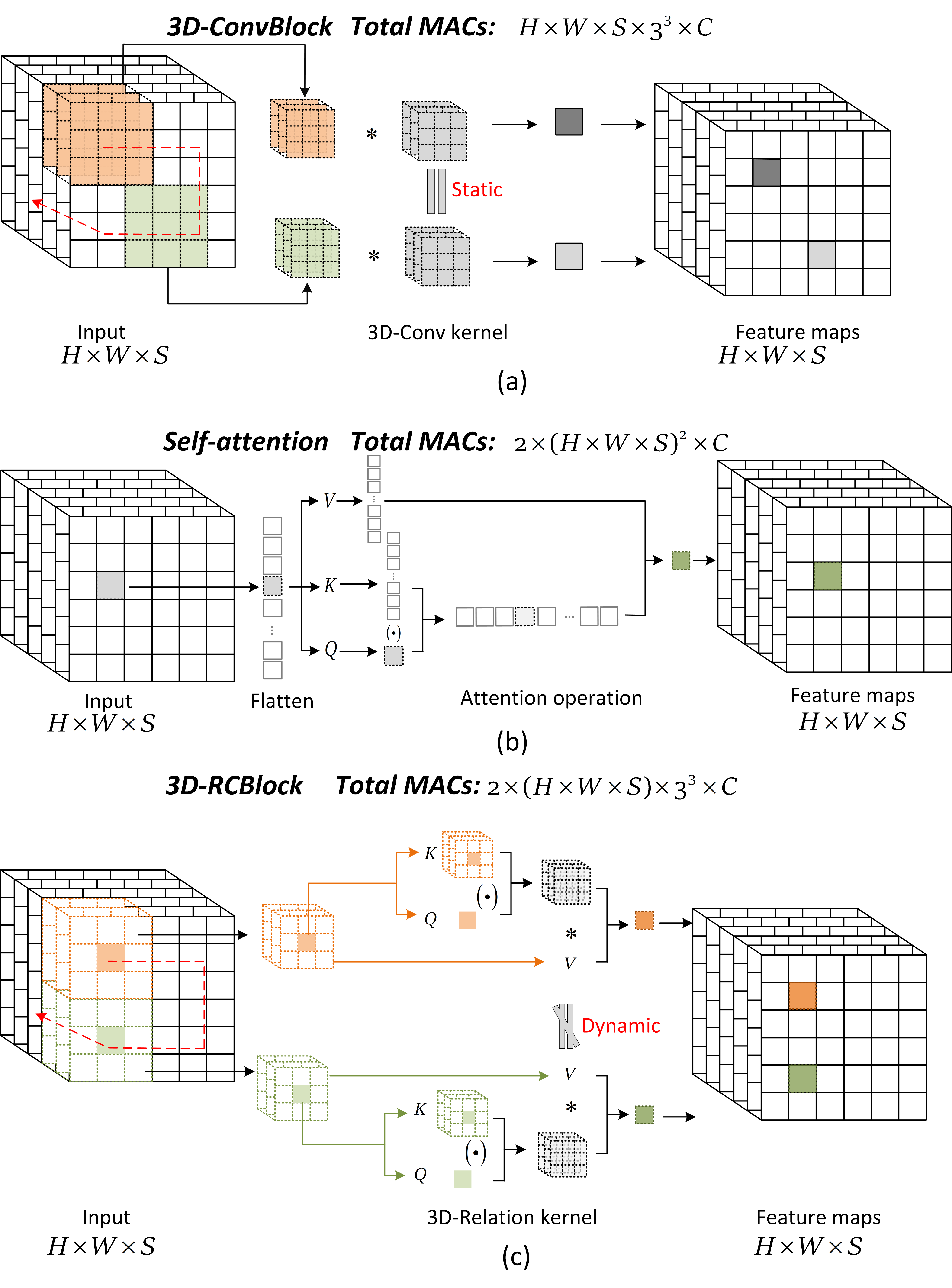}
	\caption{
		Comparison of the three methods, the total MACs required by each method with the same input. (a) is 3D-ConvBlock,(b) is Self-attention, and (c) is our proposed 3D-RCBlock.
	}
	\label{fig:3methodcompare}
\end{figure*}
\subsection{The 3D Relational Convolutional Block}
\label{sec:The Relation ConvNet block}
This section will elaborate on the design philosophy and core mechanisms of our proposed 3D-RCBlock, which combines the advantages of traditional convolution and self-attention mechanisms. We will use the same input data to compare the effectiveness of traditional convolution, self-attention mechanisms, and 3D-RCBlock in feature extraction, and will present Multiply-Accumulate Counts (MACs) for these three approaches. \Cref{fig:3methodcompare} displays the feature extraction processes utilized by these three modules. The input is denoted as \textbf{I} and the output as \textbf{O}, with H and W indicating the height and width of the input data space, and S representing the number of spectral bands. The design specifics and benefits of our 3D-RCBlock are detailed through its data feature extraction process.\\
\textbf{Convolutional:} Convolutional aggregates information by considering the local domain of each element of the input features. For convolution, sequential feature extraction using overlapping windows for input \textbf{I} enables effective extraction of local features and the parameters of the convolutional layer are shared over the entire image, reducing the overall number of parameters. Most of all, the translation invariance of convolutional can enable the model to process images at different locations and better generalise to new data, which can be expressed as follows:
\begin{equation}
    O(i, j, l) = \sum_m \sum_n \sum_zI(i+m, j+n, l+z) \cdot K(m, n, z)
\end{equation}

Where \textbf{O} is the output feature map, \textbf{I }is the input image or feature map, and K is the convolutional kernel. However, this reuse of the same convolutional kernel parameters at different locations prevents the convolutional from performing adaptive extraction of features.

As shown in \Cref{fig:3methodcompare}(a), when 3D ConvNet is used to process HSI data, the input \textbf{I} is H×W×S, the depthwise convolutional kernel is 3×3×3. And the output of the convolutional is also H×W×S, the MACs for depthwise 3D convolutional operation is $H \times W \times S \times 3^3\times C$. \\
\textbf{Self-attention:} In self-attention, three learnable weight matrices, $W^{Q}$, $W^{K}$, and $W^{V}$, are predefined to facilitate the learning of multiple meanings associated with Query, Key, and Value. Each input token is linearly mapped through these matrices to produce the query matrix Q, key matrix K, and value matrix V. The attention score is calculated using the dot product between the query matrix Q and key matrix K, while the score weights are derived through the softmax function. This function models the relationships within the image, enabling effective capture of dependencies of different pixels and enhancing image understanding. The self-attention mechanism then aggregates embeddings by computing a weighted sum, with weights based on the pairwise relationships among the embeddings, which can be expressed as follows:
\begin{equation}
    O(i, j, l)  = \text{softmax}(\frac{q_{ijl} K^T}{\sqrt{d_k}}) V
\end{equation}
Where $q_{ijl}$ is a row of the query matrix representing a particular query vector. $q_{ijl}$ is used in the transpose $K^T$ dot product operation on the key matrix to compute the similarity of $q_{ijl}$ to all keys. $K$ and $V$ are flatten features, each of which consists of $N=H\times W\times S$ feature vectors, shown as follows:

\begin{equation}
 K^T = \begin{bmatrix}
k_{11}  k_{21}  \cdots  k_{N1} \\
k_{12}  k_{22}  \cdots  k_{N2} \\
\vdots  \vdots  \ddots  \vdots \\
k_{1d_k}  k_{2d_k}  \cdots  k_{Nd_k}
\end{bmatrix}, \\ V = \begin{bmatrix}
v_{11}  v_{12}  \cdots  v_{Nd_v} \\
v_{21}  v_{22}  \cdots  v_{Nd_v} \\
\vdots  \vdots  \ddots  \vdots \\
v_{N1}  v_{N2}  \cdots  v_{Nd_v}
\end{bmatrix}  
\end{equation}
where $d$ is the dimension of attention heads. Generally $C=n_{head}d$, and $n_{head}$ is the number of attention heads. The corresponding theoretical MACs is   $2 \times (H \times W \times S)^2\times C$. It can be seen that self-attention does not require a specific setting for the resolution, is highly flexible. However, for high-resolution inputs, self-attention is computational cost because its computational complexity is proportional to the square of the input resolution.\\
\textbf{3D-RCBlock:} By analysing convolutional and self-attention, we design an 3D-RCBlock that combines the advantages of both to enable better extraction of feature information on 3D data and reduce its computational effort. Specifically, for the input feature I, we use the same sliding overlapping window as convolutional for data extraction, with the window size set to 3.  We perform attention operations within the extracted window, but unlike self-attention, we use the centre pixel within the window as  query matrix Q,  which can be expressed as follows:

\begin{equation}
\begin{split}
O(i, j, l) = \sum_m \sum_n \sum_z(\frac{e^{-(q_{ijl}+K_(i-m)(j-n)(l-z)}}{Norm} ) \\\times V_(i-m)(j-n)(l-z) 
\end{split}
\end{equation}
\begin{flalign}
Norm = \sum_m \sum_n \sum_ze^{-(q_{ijl}+K_(i-m)(j-n)(l-z))}&
\end{flalign}
where m, n, and z represent the computation of the attention weights on three independent dimensions with the corresponding values $V_(i-m)(j-n)(l-z)$. $e^{-(q_{ijl}+K_(i-m)(j-n)(l-z))}$ is an exponential weight where $q_{ijl}$ is the query vector and $K_(i-m)(j-n)(l-z)$ is the key vector, which is represented by their sum to indicate their similarity at a specific offset. And $Norm$ is a normalisation factor that ensures that the weights of all the values of V used for weighting sum to one. As shown in \Cref{fig:3methodcompare}(c) , the proposed 3D-RCBlock generates dynamic kernels for the input cube according to the relationship among pixels in this cube.   The MACs of the 3D-RCBlock is $ 2\times(H \times W \times S) \times 3^3 \times C$. \Cref{tab:3D-SA-RC} lists the MACs for three operations when process 3D HSI data. Although the computational load of the 3D-RCBlock is proportional to the input resolution, the reduction in computational load is quite significant compared to the MACs of self-attention, and the computational load is greatly reduced in high-resolution scenarios.

\begin{table}[t]\label{tab:table6}
	\centering

\caption{Theoretical Multiply-Accumulate Counts (MACs) of Convolutional, Self-attention and 3D-RCBlock.}
\label{tab:3D-SA-RC}
\resizebox{\columnwidth}{!}{
\tiny
\begin{tabular}{>{\centering\arraybackslash}p{1.5cm}ccl} \hline  
			   Operation&\multicolumn{3}{c}{Theoretical MACs}\\\hline
 Convolutional& \multicolumn{3}{c}{$ H \times W \times S \times 3^3 \times C$}\\
 Self-attention& \multicolumn{3}{c}{$2 \times (H \times W \times S)^2 \times C$}\\ 
			    3D-RCBlock&\multicolumn{3}{c}{$ 2\times(H \times W \times S) \times 3^3 \times C$}\\ \hline 
\end{tabular}}
\end{table}
In our 3D-RCBlock, we do the attention operation within a 3 × 3 × 3 window by a convolution-like operation, which greatly reduces the amount of operations. Our 3D-RCBlock performs the attention operation within the window, which makes our 3D-RCBlock also has the translation invariance of convolutional network and has the same flexibility as self-attention. The 3D-RCBlock combines the advantages of convolutional and self-attention, which reduces the computation amount and at the same time, it can effectively extract the deeper features of the HSI data. Unlike traditional static convolutional kernels, the 3D-RCBlock utilizes dynamic convolutional kernels inspired by the self-attention mechanism, which adapt based on the distribution characteristics of the data. This has been demonstrated experimentally in \Cref{sec:Dynamic Kernel}.

\section{Experiments}
\label{sec:experiments}
In this section, we conduct a comparative analysis of 3D-RCNet against five HSI classification methods known for their excellent performance: 3D CNN, LWNet, SSFTT, SpectralFormer, and GraphGST. These five comparison methods were selected for analysis because 3D CNN and LWNet are classic and well-performing methods based on 3D CNN, proven robust in capturing spatial features in high-dimensional data. On the other hand, SSFTT, SpectralFormer, and GraphGST are effective ViT-based methods known for their ability to capture long-range dependencies and complex patterns in data. Choosing these methods for comparative experiments allows for a better assessment of 3D-RCNet's performance, highlighting its strengths and limitations in HSI classification. The comparison utilized three metrics commonly employed in HSI classification tasks: overall accuracy (OA), average accuracy (AA), and the Kappa coefficient. These metrics collectively provide a comprehensive and reliable assessment of the classification capabilities of each model across various classes and datasets.

\begin{table*}[t]\label{tab:table0}
	\centering
\caption{Sample distribution information of the datasets}
\label{tab:distribution of dataset}
\fontsize{18}{24}\selectfont
\setlength{\tabcolsep}{2pt}
\resizebox{\textwidth}{!}{
		\begin{tabular}{c|c >{\centering\arraybackslash}p{2.5cm}>{\centering\arraybackslash}p{3.5cm}|c >{\centering\arraybackslash}p{2.5cm}>{\centering\arraybackslash}p{3.5cm}|c >{\centering\arraybackslash}p{2.5cm}>{\centering\arraybackslash}p{3.5cm}}\hline 
 \multicolumn{4}{c}{Indian Pines Dataset}& \multicolumn{3}{c}{Pavia University Dataset}& \multicolumn{3}{c}{Houston 2013 Dataset}\\ \hline  
			   No.&Land Cover Type& \# of Samples&\# of training Samples& Land Cover Type& \# of Samples&\# of training Samples&Land Cover Type& \# of Samples&\# of training Samples\\ \hline    
			     1&Alfalfa&46
 &10& Asphalt& 6631
 &150&Halthy Grass& 1251 &150\\ 
			    2&Corn-notill&1428
 &150& Meadows& 18649
 &150&Stressed Grass& 1254 &150\\  
			    3&Corn-mintill&830
 &150& Gravel& 2099
 &150&Synthetic Grass& 679 &150\\
			    4&Corn&237
 &10& Trees& 3064
 &150&Trees& 1244 &150\\
			    5&Grass-pasture&483
 &150& Painted metal sheets& 1345
 &150&Soil& 1242 &150\\  
			    6&Grass-trees&730
 &150& Bare Soil& 5029
 &150&Water& 325 &150\\
			    7&Grass-pasture-mowed&28
 &10& Bitumen& 1330
 &150&Rsidential& 1268 &150\\
 8& Hay-windrowed& 478
 &150& Self-Blocking Bricks& 3682
 &150&Commercial& 1244 &150\\
 9& Oats& 20
 &10& Shadows& 947
 &150&Road& 1252 &150\\
 10& Soybean-notill& 972
 &150& -& - &- &Highway& 1227 &150\\
 11& Soybean-mintill& 2455
 &150& -& - &- &Railway& 1235 &150\\
 12& Soybean-clean& 593
 &150& -& - &- &Parking Lot 1& 1233 &150\\
 13& Wheat& 205
 &10& -& - &- &Parking Lot 2& 469 &150\\
 14& Woods& 1265
 &150& -& - &- &Tennis Court& 428 &150\\
 15& Buildings-Grass-Trees-Drives& 386
 &150& -& - &- &Running Track& 660 &150\\
 16& Stone-Steel-Towers& 93
 &10& -& - &- &-& - &- \\ \hline 
 & Total& 10249
 &1560& Total& 42776 &1350&Total& 15029 &2250\\ \hline 
		\end{tabular}}
\end{table*}
\begin{figure*}[htbp]
	\centering
	\includegraphics[width=0.75\textwidth]{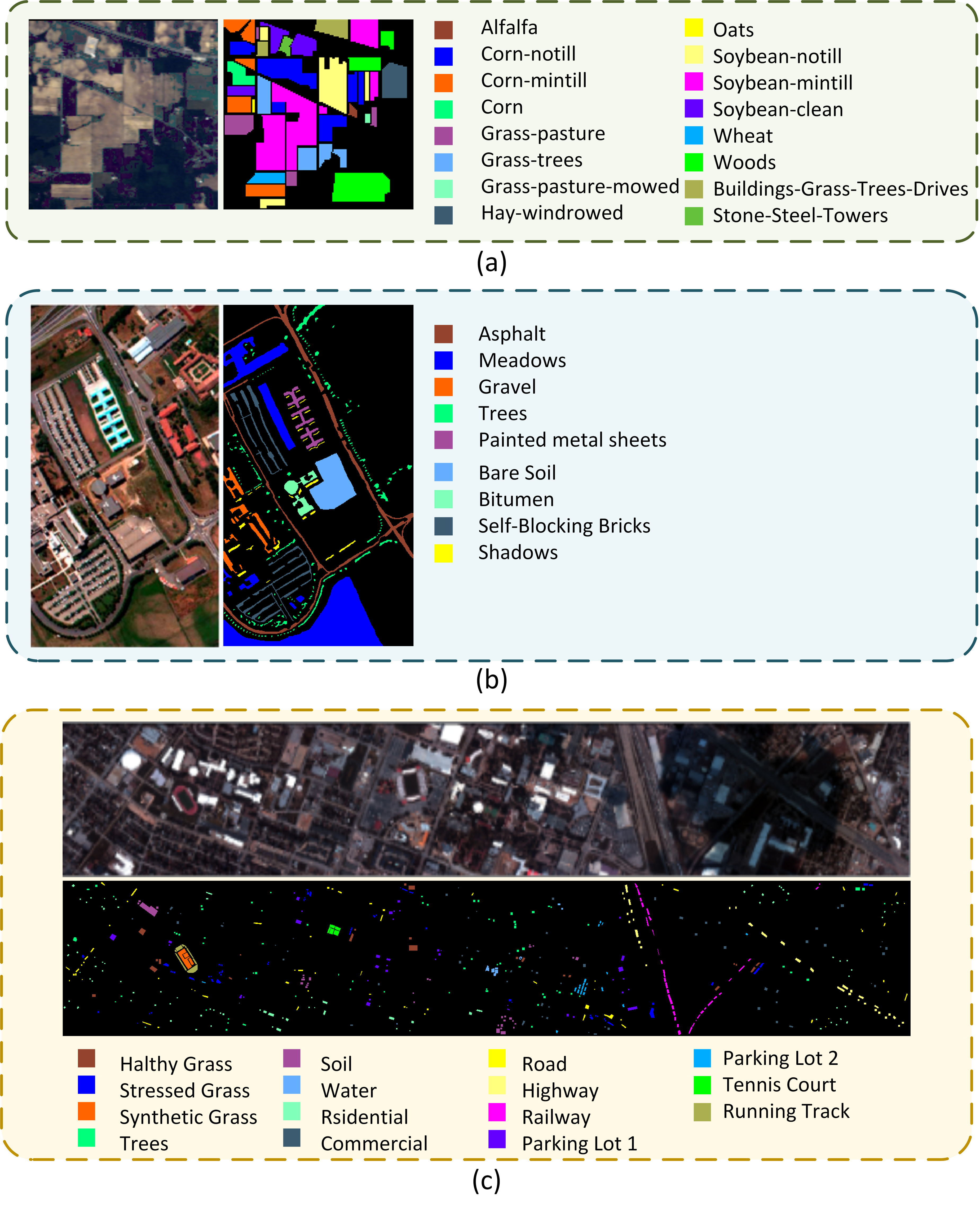}
	\caption{False color composites of experimental HSI datasets and the ground truth of land cover type. (a) Indian Pines Dataset.  (b) Pavia University Dataset. (c) Houston 2013 Dataset. }
  \label{fig:datasetshow}
\end{figure*}
\subsection{Data Description}
\label{sec:Dataset description}
The Indian Pine dataset, the Pavia University dataset, and the Houston 2013 dataset were chosen for their distinctive features and widespread use in remote sensing. The Indian Pine dataset presents varied agricultural landscapes, the Pavia University dataset includes detailed urban settings, and the Houston 2013 dataset offers diverse urban and semi-urban scenes with high-resolution data, ideal for testing the granularity of classification methods. These datasets ensure a comprehensive evaluation across a wide range of real-world conditions, effectively testing the robustness and accuracy of the methods in diverse and challenging environments. \Cref{fig:datasetshow} displays the false color composites and the ground truth of these three datasets. A detailed description of these datasets follows.\\
\textbf{Indian Pines Dataset:} Hyperspectral remote sensing images captured by the Airborne Visible/Infrared Imaging Spectrometer (AVIRIS) over northwestern Indiana, USA, in 1992. This dataset features a 145x145 pixel grid with 220 spectral bands, covering a wavelength range from 0.4µm to 2.5µm. Designed primarily for agricultural and forestry applications, each pixel in the dataset captures a rich series of reflectance values across these bands. This detailed spectral information makes the dataset highly valuable for analyzing and distinguishing between various natural and anthropogenic features, such as farmlands, forests, and built-up areas. Although the dataset originally included all 220 bands, analyses often exclude certain bands that fall within the water absorption regions, typically using around 200 bands for conducting detailed land cover classifications into 16 distinct categories. The Indian Pines Dataset is widely utilized in remote sensing research for developing and testing algorithms for hyperspectral image processing and classification.\\
\textbf{Pavia University Dataset:} Hyperspectral remote sensing image data acquired by the Reflectance Optical System Imaging Spectroradiometer (ROSIS) sensor from the urban area near the University of Pavia, Italy, in 2001. The dataset contains multispectral images with a total of 610x340 pixels, each pixel having information in 115 spectral bands with wavelengths ranging from 0.43 µm to 0.86 µm.These bands represent different electromagnetic wavelengths used to capture a variety of features on the surface. The image contains many different features of the urban area such as buildings, roads, grass, trees, etc. After removing 12 noise bands, the remaining 103 bands are used for 9 land cover classifications.\\
\textbf{Houston 2013 Dataset:} Hyperspectral airborne remote sensing imagery acquired over the University of Houston campus and surrounding urban areas by the ITRESCASI 1500 hyperspectral images, Houston, Texas, USA, in 2012, was initially used in the 2013 IEEE GRSS data fusion competition. The dataset contains 349x1905 pixel hyperspectral images with 144 simultaneous spectral bands in the wavelength range of 0.36µm to 1.05µm for 15 land cover classifications. Since the images cover a variety of different features in urban areas such as buildings, roads, fields, water bodies, etc., this dataset is challenging to develop effective classification algorithms to differentiate between these feature types.

In our experiments, we divided the above three datasets into training and test sets. For the Pavia University and Houston 2013 dataset, we took 150 samples from each category as the training set and used the others as the test set. For the IndianPines dataset, since 6 of its 16 classes have less than 300 total samples, especially class1 (Alfalfa), class7 (Grass-pasture-mowed), and class9 (Oats) have only 46, 28, and 20 samples, respectively, for these classes with relatively small number of samples, we randomly select 10 samples as the training set and used the others as the test set. \Cref{tab:distribution of dataset} lists the types of land cover classes, the number of samples per class, and the number of samples used for training in each of the three datasets.

\subsection{Implementation Details}
\label{sec:Experiemtns Settings}
The experiments were conducted using a high-specification server equipped with a 2.10GHz Intel(R) Xeon(R) Gold 6230 CPU, 512 GB RAM, and  Nvidia Tesla V100 32 GB graphics cards. The proposed 3D-RCNet model was implemented using the Pytorch 1.12.0 open source framework. The model training was performed exclusively on the Nvidia Tesla V100 32 GB graphics card.

For the training setup, the model processed three different datasets, maintaining the same network architecture across all. Each dataset was uniformly cropped into 3D cubes with a spatial resolution of 27x27 for input, while the spectral dimensions were left unaltered. The batch size was set at 64 for all datasets with the training running for 300 epochs. The optimization was done using AdamW, with a learning rate of 5e-4 and a weight decay of 1e-5. The learning rate scheduling was managed by a cosine annealing strategy, beginning with a warm-up phase covering the first 30 epochs where the learning rate increased from 10\% of its initial value. For the remaining 270 epochs, the learning rate decayed from 1e-5 to 5e-6.
\begin{figure*}[t]
	\centering
	\includegraphics[width=0.75\textwidth]{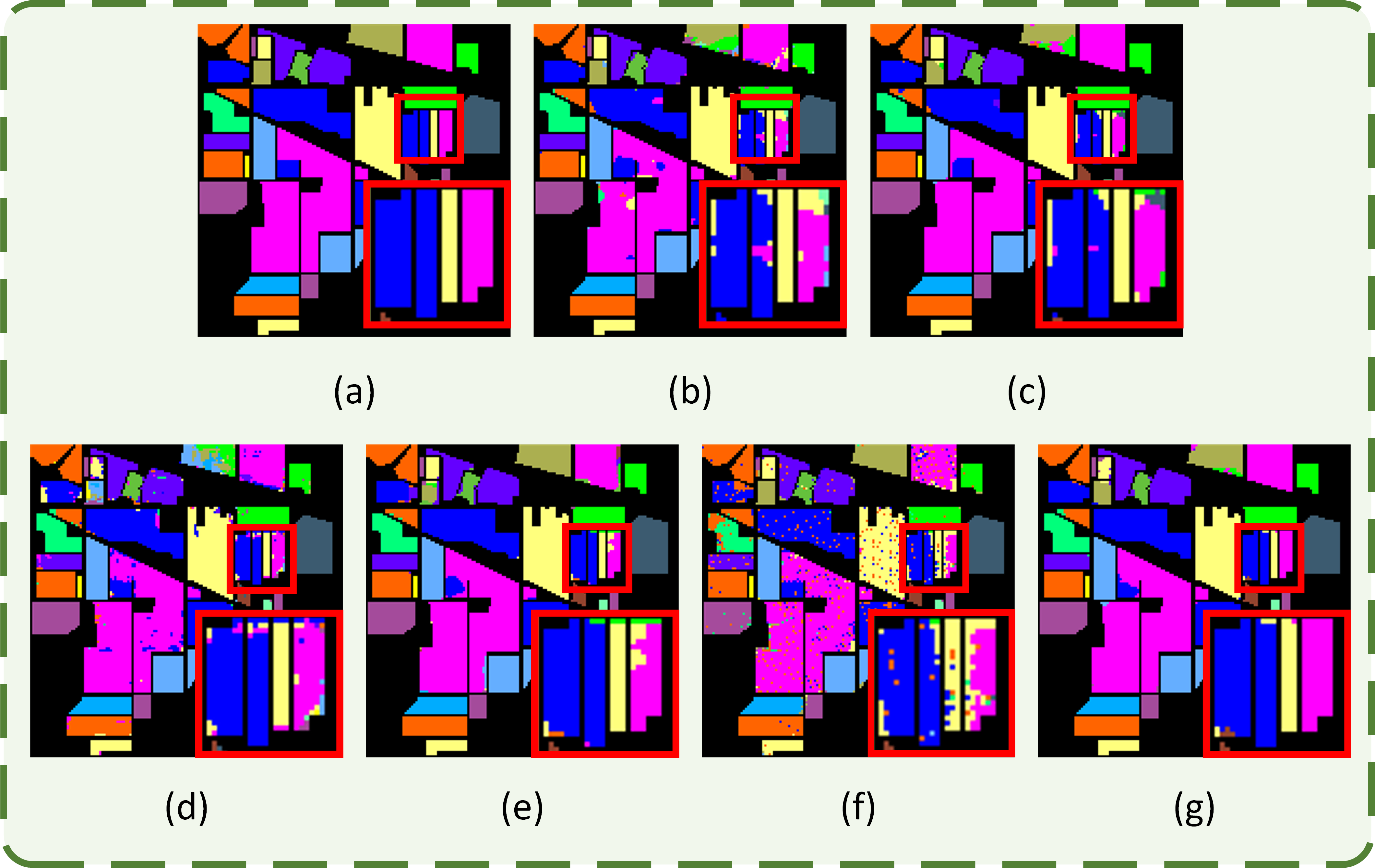}
	\caption{In comparative experiments conducted on the Indian Pine dataset, we visualize the prediction results. (a) is the Ground Truth. (b) is the prediction result of 3D CNN. (c) is the prediction result of LWNet. (d) is the prediction result of SSFTT. (e) is the prediction result of SpectralForm. (f) is the prediction result of GraphGST. (g) is the prediction result of our proposed 3D-RCNet. }
  \label{fig:VR1}
\end{figure*}

\subsection{Comparison Experiments}
\label{sec:Comparison Experiments}
In the Indian Pine dataset, 3D-RCNet's classification performance significantly surpasses five other methods across 16 land cover categories, with the relevant data listed in \Cref{tab:IndianPine Datasets} and the corresponding visual comparison results shown in \Cref{fig:VR1}. \Cref{tab:IndianPine Datasets} indicates that 3D-RCNet achieved an OA of 98.50\%, AA of 98.03\%, and a Kappa of 98.26\%. In these three key performance indicators, 3D-RCNet outperforms the other five comparison methods. Notably, although SSFTT demonstrates excellent performance due to its superior deep semantic feature extraction capabilities, surpassing several other advanced methods, 3D-RCNet's OA is still 1.32\% higher than SSFTT. 

\begin{table}[t]\label{tab:table1}
	\centering
\caption{Comparison experimental results on Indian Pine dataset}
\label{tab:IndianPine Datasets}
\fontsize{9}{12}\selectfont
\resizebox{\columnwidth}{!}{
\begin{tabular}{c|c c|c c c|c} 
\hline
Models&3D CNN& LWNet& SSFTT& SpectralFormer &GraphGST&3D-RCNet\\ 
\hline    
    1&91.67 & 97.22& 100.00& 77.78         &100.00  &100.00
\\ 
    2&94.84 & 96.01&  94.99& 90.06         &94.91   &97.18
\\  
    3&99.12 & 98.82&  99.12& 94.26         &99.26   &98.82
\\
    4&100.00&100.00& 100.00& 98.85         &74.89   &100.00
\\
    5&100.00&100.00& 100.00& 99.10         &96.99   &100.00
\\  
    6&100.00&99.83& 99.83& 98.97           &97.76   &100.00
\\
    7&100.00&100.00& 100.00& 94.44         &94.44   &100.00
\\
    8& 100.00& 100.00& 100.00& 100.00      &100.00  &100.00
\\
    9& 100.00& 100.00& 100.00& 100.00      &100.00  &100.00
\\
    10& 99.39& 100.00& 99.51& 96.35        &99.15   &98.66
\\
    11& 92.41& 98.39& 95.05& 90.54         &93.02  &99.39
\\
    12& 98.42& 99.77& 99.55& 96.61         &96.61  &99.55
\\
    13& 100.00 & 100.00& 100.00& 100.00    &97.95  &100.00
\\
    14& 98.92  & 99.91& 99.82& 98.92       &99.46  &99.82
\\
    15& 66.49  & 70.74& 89.63& 68.35       &100.00 &88.30
\\
    16& 90.36  & 91.57& 83.13& 84.33       &97.59  &86.75  
\\ \hline 
 OA(\%)& 95.23 & 97.46& 97.12& 93.08       &96.02&\textbf{98.50}
\\
 AA(\%)& 95.73 & 97.02& 97.54& 93.04       &96.38&\textbf{98.03}
\\
 KAPPA(\%)& 94.47& 97.05& 96.66& 91.97     &95.41&\textbf{98.26}\\ \hline 
		\end{tabular}}
\end{table}
Furthermore, the visual comparison results of 3D-RCNet on the Indian Pine dataset are displayed in \Cref{fig:VR1}(g), where the bottom right corner provides an enlarged view for easier visual comparison. Other methods exhibit varying degrees of classification errors, such as the salt-and-pepper noise-like visual effects shown in \Cref{fig:VR1}(f), but the proposed 3D-RCNet's classification results are noticeably superior to the comparison group methods, much closer to the ground truth. 

\begin{table}[t]\label{tab:table2}
	\centering
 
\caption{Comparison experimental results on Pavia University dataset}
\label{tab:Pavia University Datasets}
\fontsize{9}{12}\selectfont
\resizebox{\columnwidth}{!}{
		\begin{tabular}{c |c c |c c c|c} \hline  
			   Models&3D CNN& LWNet& SSFTT& SpectralFormer &GraphGST&3D-RCNet\\ \hline    
			     1&91.09&96.06& 94.69& 83.13 &98.06&98.87
\\ 
			    2&92.05&95.69& 98.74& 92.83 &98.03&98.64
\\  
			    3&92.45&95.95& 98.50& 89.34 &92.82&100.00
\\
			    4&98.55&98.68& 95.05& 97.60 &99.11&97.74
\\
			    5&99.52&99.92& 97.43& 100.00 &100.00&100.00
\\  
			    6&93.14&98.09& 99.96& 90.32 &99.92&100.00
\\
			    7&97.48&99.76& 100.00& 96.99 &100.00&100.00
\\
                    8& 95.20& 99.11& 99.08& 92.63 &95.95&98.98
\\
                    9& 97.28& 98.82& 99.88& 99.65 &99.74&99.25
\\ \hline 
        OA(\%)& 93.26& 96.86& 98.03& 91.65 &98.06&\textbf{98.96}
\\
        AA(\%)& 95.20& 98.01& 98.15& 93.61 &98.18&\textbf{99.28}
\\
 KAPPA(\%)& 91.13& 95.84& 97.38& 89.00 &97.41&\textbf{98.61}\\ \hline 
		\end{tabular}}
\end{table}

\begin{figure*}[htbp]
	\centering
	\includegraphics[width=0.75\textwidth]{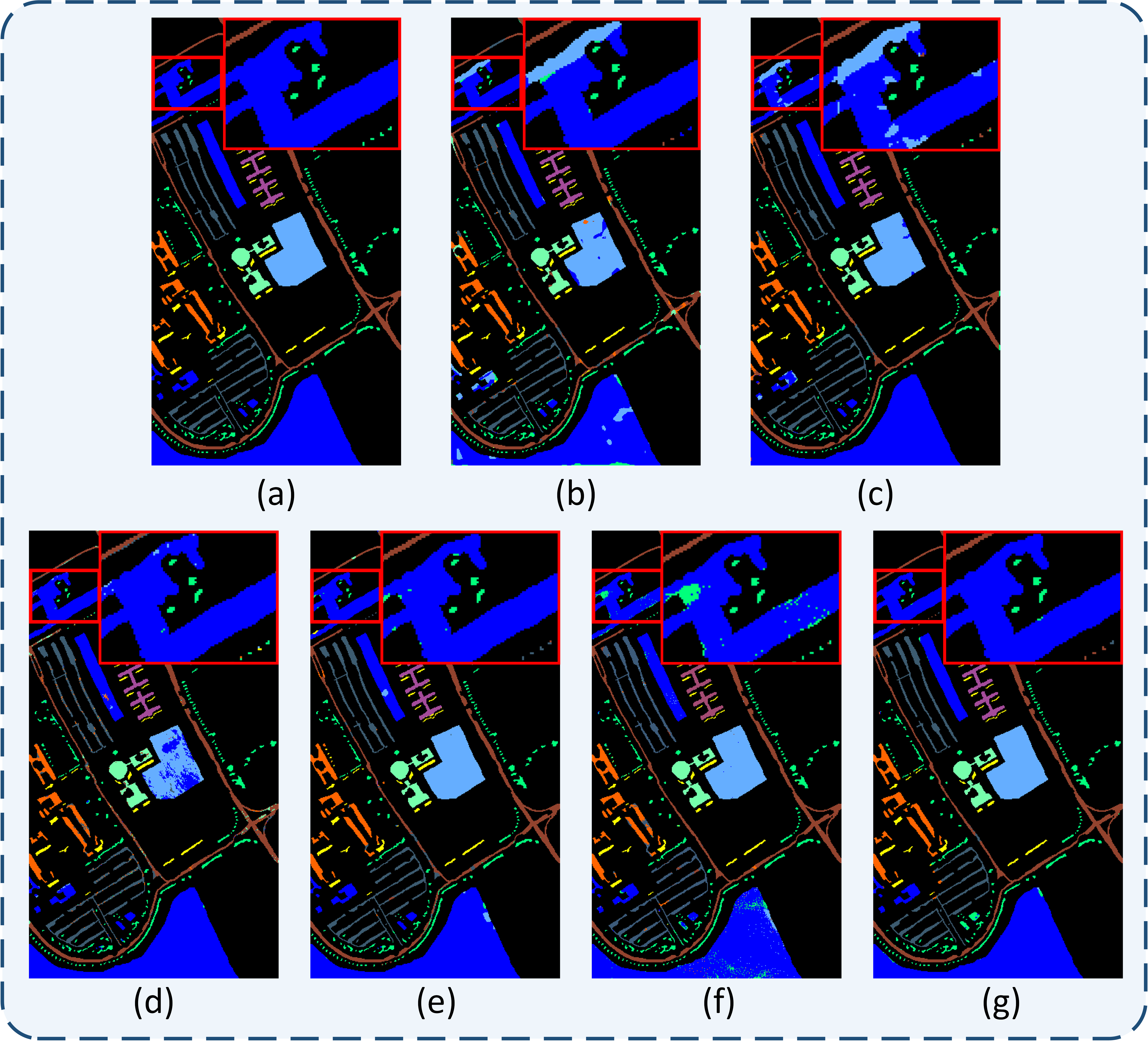}
	\caption{In comparative experiments conducted on the  Pavia University dataset, we visualize the prediction results. (a) is the Ground Truth. (b) is the prediction result of 3D CNN. (c) is the prediction result of LWNet. (d) is the prediction result of SSFTT. (e) is the prediction result of SpectralForm. (f) is the prediction result of GraphGST. (g) is the prediction result of our proposed 3D-RCNet. }
  \label{fig:VR2}
\end{figure*}

 \Cref{tab:Pavia University Datasets} lists the classification performance on the Pavia University dataset, where our proposed 3D-RCNet achieved an OA of 98.96\%, significantly higher than other advanced methods. Among the rest, the best performing ConvNet-based method was LWNet with an OA of 96.86\%, and the best performing ViT-based method was GraphGST, with an OA of 98.06\%. In comparison, 3D-RCNet exceeds these two methods by 2.1\% and 0.9\%, respectively. 

 \Cref{fig:VR2}(g) shows the visual classification results of 3D-RCNet, which has fewer misclassified points, particularly in the classification of grassland edge areas, demonstrating results close to the ground truth. Compared to the other five control group experiments, 3D-RCNet exhibits the most outstanding performance in the classification task on the Pavia University dataset.

\begin{figure*}[htbp]
	\centering
	\includegraphics[width=0.5\textwidth]{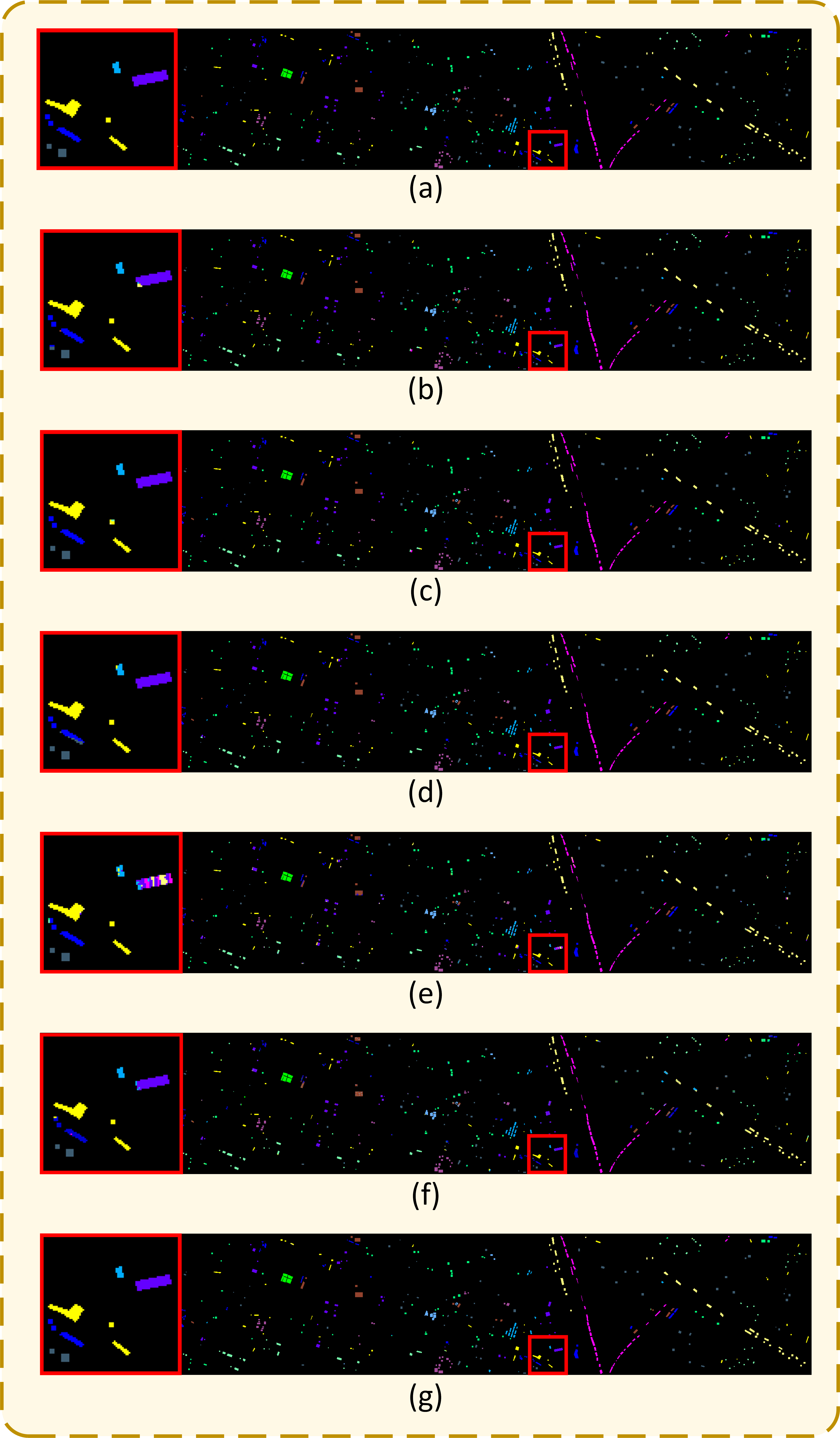}
	\caption{In comparative experiments conducted on the Houston 2013 dataset, we visualize the prediction results. (a) is the Ground Truth. (b) is the prediction result of 3D CNN. (c) is the prediction result of LWNet. (d) is the prediction result of SSFTT. (e) is the prediction result of SpectralForm. (f) is the prediction result of GraphGST. (g) is the prediction result of our proposed 3D-RCNet. }
  \label{fig:VR3}
\end{figure*}
Similarly, our proposed 3D-RCNet shows excellent performance on the Houston 2013 dataset. As shown in \Cref{tab:Houston 2013 Datasets}, 3D-RCNet has an OA of 99.23\%, an AA of 99.41\%, and a Kappa of 99.17\%, which once again outperforms all the control group methods in this set of performance metrics. As shown in \Cref{fig:VR3}, 3D-RCNet's visualisation results on the Houston 2013 dataset show effective separability, with fewer misclassified points in the zoomed-in detail section at the bottom right.

\begin{table}[t]
\label{tab:table3}
\centering
\caption{Comparison experimental results on Houston 2013 dataset}
\label{tab:Houston 2013 Datasets}
\fontsize{9}{12}\selectfont
\resizebox{\columnwidth}{!}{
		\begin{tabular}{c| c c| c c c|c} \hline  
			   Models&3D CNN& LWNet& SSFTT& SpectralFormer &GraphGST&3D-RCNet\\ \hline    
			     1&96.64&98.64& 98.00& 96.00 &87.27&98.64
\\ 
			    2&95.20&99.01& 97.10& 89.31 &98.12&100.00
\\  
			    3&99.82&100.00& 99.09& 99.09 &98.61&100.00
\\
			    4&97.62&99.54& 97.90& 86.47 &91.86&99.18
\\
			    5&99.45&100.00& 100.00& 95.88 &99.72&100.00
\\  
			    6&100.00&100.00& 100.00& 86.86 &95.80&100.00
\\
			    7&91.68&96.06& 98.66& 73.23 &89.27&98.30
\\
 8& 94.70& 96.71& 97.62& 85.56 &96.01&98.90
\\
 9& 96.82& 97.19& 99.82& 85.57 &90.08&97.28
\\
 10& 98.33& 99.44& 100.00& 91.55 &91.69&100.00
\\
 11& 97.05& 100.00& 100.00& 92.35 &98.39&100.00
\\
 12& 96.12& 98.25& 96.21& 86.98 &95.87&98.80
\\
 13& 95.61& 100.00& 99.37& 82.45 &94.74&100.00
\\
 14& 100.00& 100.00& 100.00& 100.00 &100.00
&100.00
\\
 15& 100.00& 100.00& 100.00& 97.65 &99.58&100.00
\\ \hline 
 OA(\%)& 96.58& 98.70& 98.69& 89.19 & 94.42&\textbf{99.23}
\\
 AA(\%)& 97.14& 98.99& 98.92& 89.93 &95.13&\textbf{99.41}
\\
 KAPPA(\%)& 96.29& 98.59& 98.57& 88.28 &93.94&\textbf{99.17}\\ \hline 
		\end{tabular}}
\end{table}
Overall, the comparative experiment results conducted on these three datasets for six different methods demonstrate that our proposed 3D-RCNet offers the most stable performance and optimal results. Furthermore, the results from other methods further validate the necessity of developing 3D-RCNet.
\begin{enumerate}
\item On the three datasets, the performance of the two methods based on ConvNet has shown to be stable, whereas methods based on ViT exhibit significant fluctuations due to their sensitivity to data. Nonetheless, when optimized for data, the ViT-based methods can outperform traditional ConvNet methods. This also underscores the necessity of adopting a hybrid structure in 3D-RCNet, which aims to combine the high performance of ViT-based methods with the stability of ConvNet methods.

\item We note that classical ViT-based methods such as SpectralFormer perform well when there is a sufficient amount of data, but their performance falters in experiments with limited data due to fewer training samples. In the field of hyperspectral image (HSI) classification, where the cost of labeling data is high, reducing dependence on large data volumes poses a challenge. 3D-RCNet maintains excellent performance even with fewer samples, a fact that is corroborated by our comparative experiments.

\item The experimental results also display the sensitivity of different methods to data volume. ViT-based methods are sensitive to data, which results in varying performance across different datasets. For instance, SSFTT significantly outperforms GraphGST on the Indian Pine dataset, whereas the opposite is true on the Pavia University dataset. Such variability in adaptability to data highlights the importance of maintaining consistent performance across different data qualities or volumes. 3D-RCNet is precisely such a method that can adapt to varying data requirements while maintaining stability.
\end{enumerate}

\section{Ablation Study}\label{sec:Ablation Study}
This section introduces the 3D-RCBlock, which retains the same three-dimensional convolutional window structure and translational operations as the traditional 3D ConvNet but incorporates an attention mechanism to capture long-range dependencies among patches within the 3D window. In the 3D-RCNet, traditional 3D-ConvBlock are used at shallow layers, while 3D-RCBlocks with a kernel size of 3×3 replace them in the third and fourth stages of the model to achieve optimal classification results. Additionally, multiple sets of ablation experiments confirm the effectiveness of the 3D-RCBlock in multi-stage feature extraction within the 3D-RCNet framework, thereby substantiating the rationality of the framework design. In addition, we visualise the properties of 3D-RCBlock's dynamic kernel re-feature extraction process, demonstrating the reality of 3D-RCBlock's integration of self-attention mechanisms into convolutional operations. 
%It's important to note that, in the ablation study, in order to facilitate internal comparative analysis, we did not train the model completely as in the comparative experiments. Instead, we focused on the performance of various modules within the model.\\

\subsection{Kernel Size}\label{sec:kernel size}
\begin{figure}[t]
	\centering
	\includegraphics[width=\columnwidth]{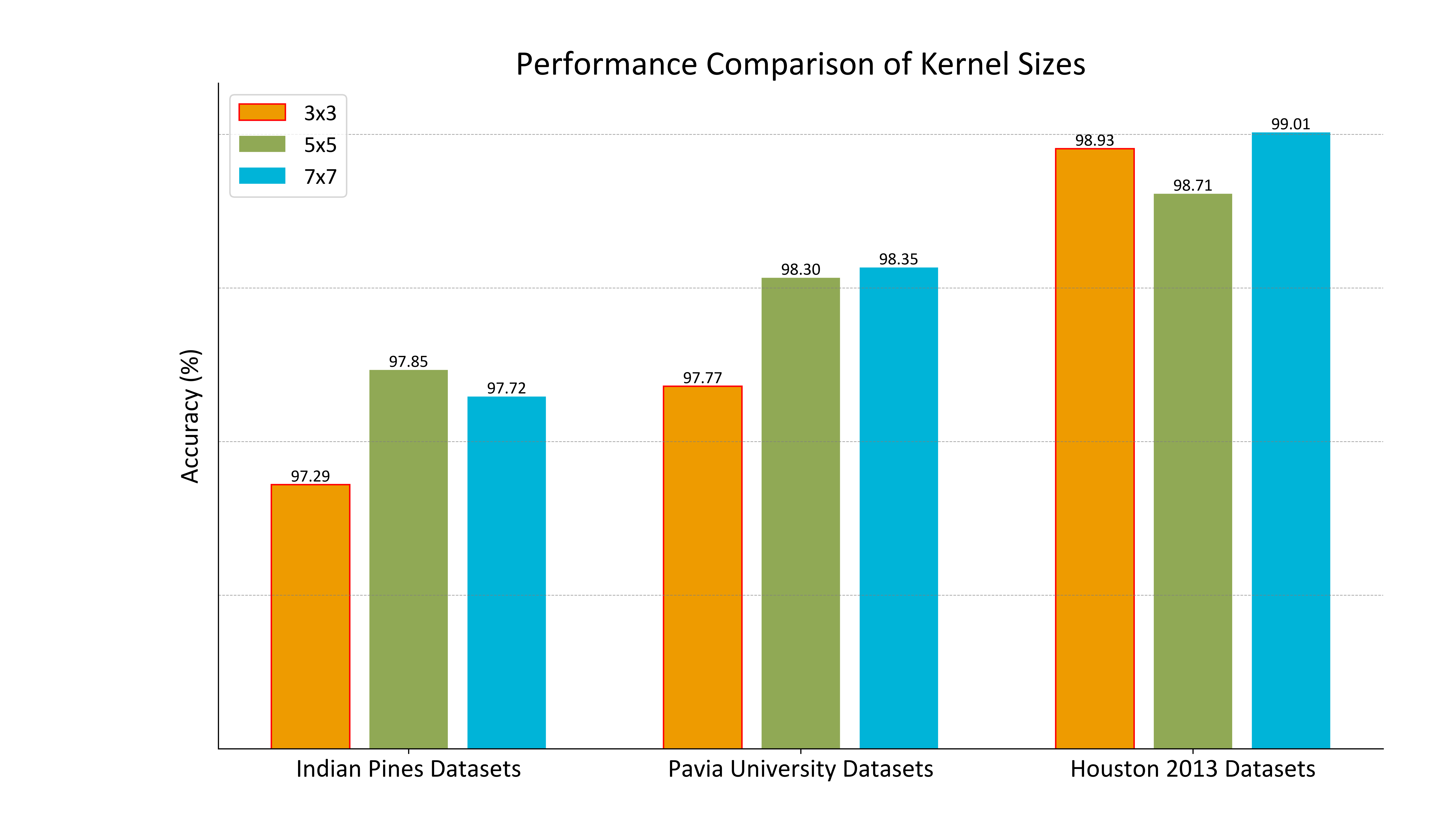}
	\caption{Experimenting the effect of different kernel sizes on classification accuracy in Stage 4. }
  \label{fig:kernelsize}
\end{figure}
In convolutional neural networks, generally speaking, the larger the size of the convolutional kernel, the larger the window size on the feature map and the better the feature extraction ability. In order to verify that different convolutional kernel sizes comply with this property in our proposed 3D-RCBlock, we set different kernel sizes to determine the most suitable convolutional kernel size for 3D-RCNet.

As shown in \Cref{fig:kernelsize}, we set the convolutional kernel size of 3D-RCBlock for Stage 4 to 3×3, 5×5,7×7 and tested the classification performance on each of the three datasets, and there is but only a small improvement in the classification accuracy as the kernel size increases. Similarly on different datasets, the degree of improvement is not the same, and even the accuracy decreases with increasing kernel size instead. This is because in Stage 4, 3D-RCBlock extracts the deep semantic information in the feature map output from the previous stage, and the kernel size of the module has little effect on the classification accuracy in the face of the deep semantic information in different data respectively. Therefore, we set the kernel size of 3D-RCBlock as 3×3 to reduce the computational overhead during operation.

\subsection{Effect of 3D-RCBlock in different stage}
\begin{table}[htbp]\label{tab:table5}
	\centering

\caption{The effect of 3D-RCBlock on classification accuracy at different stages, replacing the last block of each stage as 3D-RCBlock.} 

\label{tab:rcstage}
\resizebox{\columnwidth}{!}{
    \fontsize{5}{8}\selectfont
    \begin{tabular}{cccc} \hline  
			   Stage with 3D-RCBlock&Indian Pines& Pavia University& Houston 2013
\\ \hline    
			     NO&92.05&94.97& 96.92
\\ 
			    1&90.23&95.07& 97.30
\\  
			    2&90.92&96.93& 96.73
\\
 3& 96.47& 96.54&96.90
\\
 4& 96.51& 95.63&97.00\\
 1, 2, 3, 4& 96.48& 98.55&98.70\\ \hline 
		\end{tabular}}
\end{table}
We propose the 3D-RCBlock as a plug-and-play module. To analyze the effects of inserting this module at different stages of the model, we designed the experiments in this section, where the last module of the traditional 3D-ConvBlock at each stage is replaced with the 3D-RCBlock. This setup helps us verify the performance of the 3D-RCBlock at various stages.

Results in \Cref{tab:rcstage} show that the performance advantages of using the 3D-RCBlock in the shallow stages (Stage 1, Stage 2) of the model are not significant, especially on the Indian Pines dataset, where it even performs slightly worse than the original model. We believe that this may be due to the fact that the primary task in the shallow stages of the model is to map data space to feature space. Using the highly data-sensitive 3D-RCBlock at this stage may hinder the model from extracting basic semantic features from the raw data.

Moreover, when the 3D-RCBlock is used in the deeper stages (Stage 3, Stage 4) of the model, there is a significant improvement in performance. This improvement may be due to the richness of the feature space in the deep stages, which contains more in-depth semantic features, requiring a more flexible module for further refined extraction. This strategy of using more flexible modules in the deeper layers of the model aligns with current advanced research, which typically uses traditional ConvNet in the shallow layers and structures like ViT in the deeper stages\cite{Graham2021}\cite{Chen2021}\cite{ZhangFCA2022}. We further validated that inserting the 3D-RCBlock in all four stages of the model can achieve superior performance. However, when designing the 3D-RCNet, we also considered the overall simplicity of the system and the convenience of deployment. Therefore, in the proposed 3D-RCNet, we still chose the same pattern as current advanced research, using the 3D-RCBlock in the deeper layers (Stage 3, Stage 4), to strike a balance between the performance and practicality of the model.

\begin{figure}[t]
	\centering
	\includegraphics[width=\columnwidth]{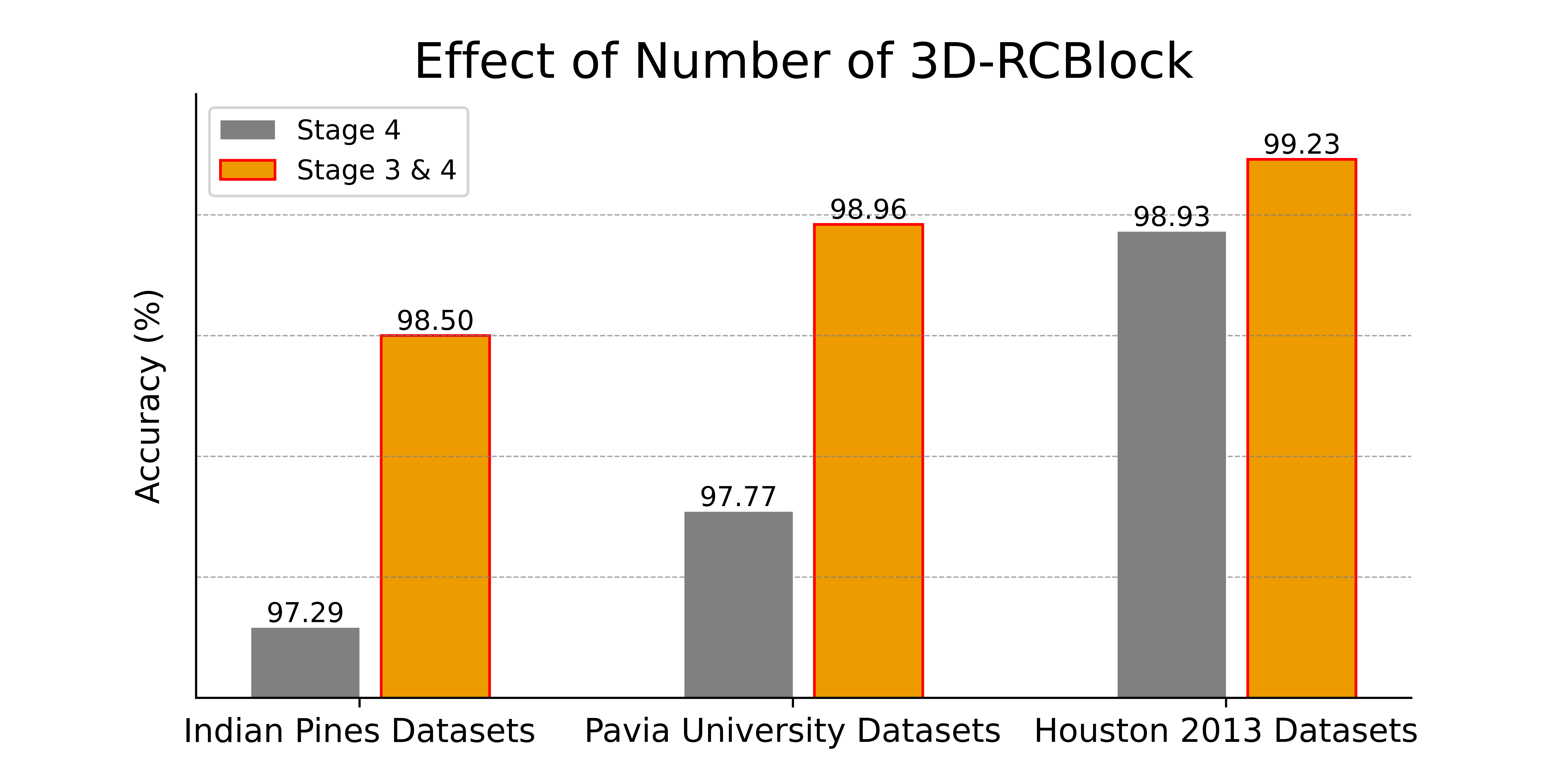}
	\caption{The effect of the number of 3D-RCBlock on classification accuracy by replacing all blocks in the stage with 3D-RCBlock. }
  \label{fig:AllRC}
\end{figure}
The performance of using 3D-RCBlock in the deeper layers was verified through experiments, but the number of 3D-RCBlocks to be used in model design still requires analysis. As shown in \Cref{tab:rcstage}, using 3D-RCBlock in Stage 4 shows better performance than in Stage 3. Initially, we replaced all blocks in Stage 4 with 3D-RCBlock and achieved good performance, as shown in \Cref{fig:AllRC}. Then, we replaced all blocks in both Stage 4 and Stage 3 with 3D-RCBlock, and found that replacing all blocks in both stages performed better than replacing only in Stage 4. Therefore, in the proposed 3D-RCNet, we decided to construct all blocks in both Stage 3 and Stage 4 using 3D-RCBlock, to fully utilize its flexibility and performance in capturing deep semantic information.

\subsection{3D-RCBlock Kernel Visualisation}\label{sec:Dynamic Kernel}
\begin{figure}[htbp]
	\centering
	\includegraphics[width=\columnwidth]{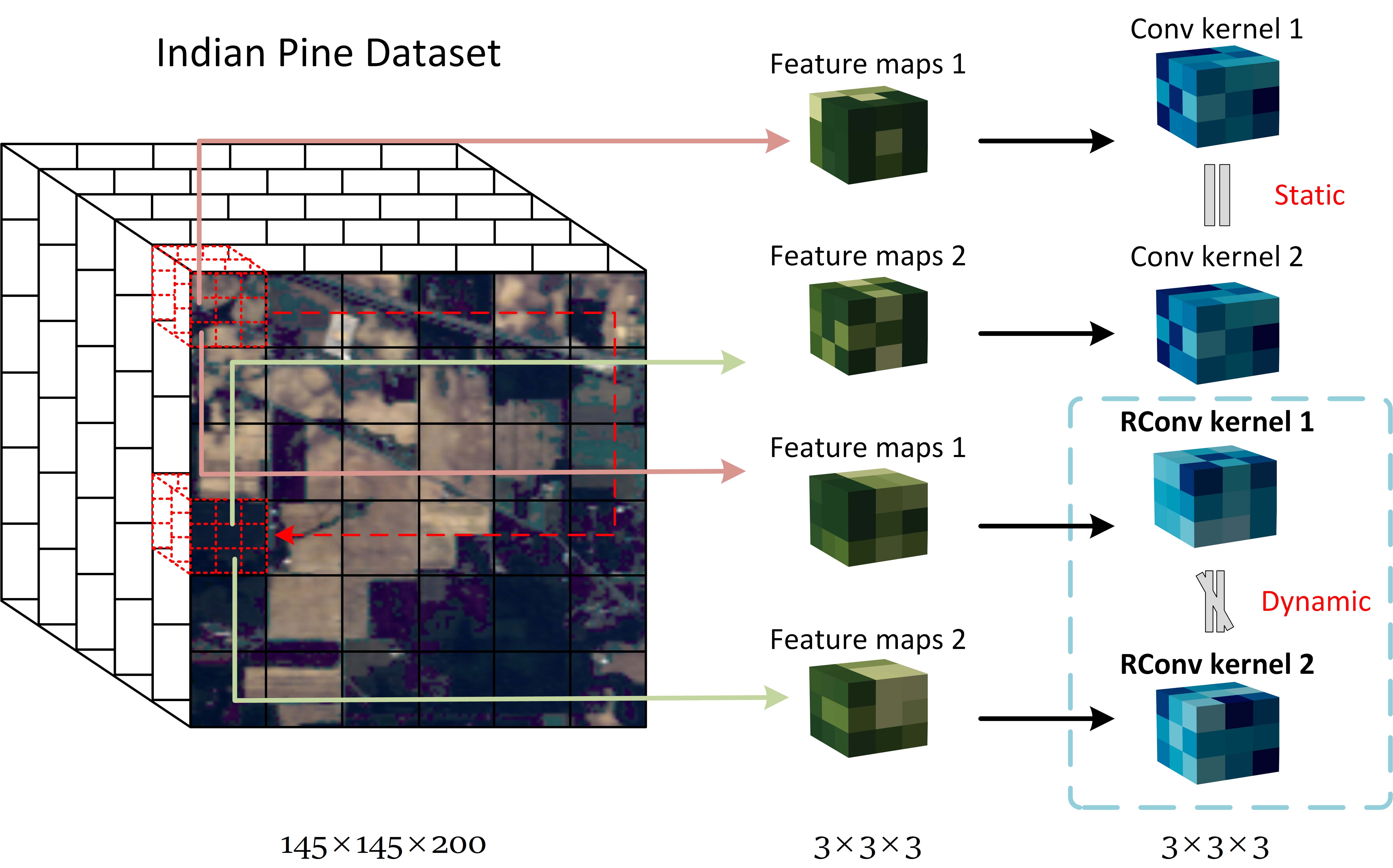}
	\caption{Feature extraction on Indian Pine dataset using 3D-ConvBlock and 3D-RCBlock: 3D-ConvBlock utilizes a static kernel to extract features from data with varying spatial positions, while our 3D-RCBlock generates dynamic kernels based on the local relationships within the data window for feature extraction. }
  \label{fig:3DvsRC}
\end{figure}
To verify that the convolutional kernels of the 3D-RCBlock differ from traditional static convolutional kernels, we conducted feature extraction using both 3D-ConvBlock and 3D-RCBlock on the Indian Pine dataset. In the experiments, we randomly selected two sets of convolution steps to examine the state of the convolutional kernels used by the 3D-ConvBlock and 3D-RCBlock and the output feature maps at these steps. As shown in \Cref{fig:3DvsRC}, the state of both types of kernels during different operations is depicted using heat maps. The results clearly show that the 3D-ConvBlock’s kernels remain static throughout the data processing, whereas the 3D-RCBlock’s kernels dynamically adapt and vary based on the data distribution extracted by the convolution window at different steps. This dynamic nature of the kernels not only effectively extracts data features but also demonstrates flexibility in adapting to various data distributions.\\

\section{Conclusion}
\label{sec:Conclusion and discussion}
In the field of HSI classification, the newly developed 3D-RCNet model significantly enhances the capability of traditional methods by seamlessly integrating the innovative 3D relational convolutional operation. This operation, which merges the convolutional strength of ConvNet with the flexibility of the Vision Transformer's self-attention mechanism, allows for a seamless adoption within existing ConvNet-based HSI classification frameworks. Not only does this integration enhance computational efficiency and reduce dependency on large training datasets, but it also facilitates a plug-and-play application, broadening the scope of the 3D-RCNet to include a variety of HSI datasets. Empirical evaluations confirm that our approach provides substantial improvements over existing ConvNet and ViT-based methods, demonstrating both robustness and versatility in practical applications. 3D-RCNet demonstrates significant advantages over existing methods in terms of theoretical MACs. Future work should focus on further reducing its computational resource requirements to enhance its applicability and effectiveness in resource-constrained real-world scenarios.

\bibliography{mybibfile}

\begin{thebibliography}{10}
\providecommand{\url}[1]{#1}
\csname url@samestyle\endcsname
\providecommand{\newblock}{\relax}
\providecommand{\bibinfo}[2]{#2}
\providecommand{\BIBentrySTDinterwordspacing}{\spaceskip=0pt\relax}
\providecommand{\BIBentryALTinterwordstretchfactor}{4}
\providecommand{\BIBentryALTinterwordspacing}{\spaceskip=\fontdimen2\font plus
\BIBentryALTinterwordstretchfactor\fontdimen3\font minus \fontdimen4\font\relax}
\providecommand{\BIBforeignlanguage}[2]{{%
\expandafter\ifx\csname l@#1\endcsname\relax
\typeout{** WARNING: IEEEtran.bst: No hyphenation pattern has been}%
\typeout{** loaded for the language `#1'. Using the pattern for}%
\typeout{** the default language instead.}%
\else
\language=\csname l@#1\endcsname
\fi
#2}}
\providecommand{\BIBdecl}{\relax}
\BIBdecl

\bibitem{2012Multi}
F.~V.~D. Meer, H.~V.~D. Werff, F.~V. Ruitenbeek, C.~A. Hecker, and T.~Woldai, ``Multi- and hyperspectral geologic remote sensing: A review grsg member news,'' in \emph{Geological Society of London, Geological Remote Sensing Group}, 2012.

\bibitem{2021HSIg}
S.~Peyghambari and Y.~Zhang, ``Hyperspectral remote sensing in lithological mapping, mineral exploration, and environmental geology: an updated review,'' \emph{Journal of Applied Remote Sensing}, vol.~15, no.~3, pp. 031\,501--031\,501, 2021.

\bibitem{Guha2020}
\BIBentryALTinterwordspacing
A.~Guha, \emph{\BIBforeignlanguage{en-US}{Mineral exploration using hyperspectral data}}, Jan 2020, p. 293–318. [Online]. Available: \url{http://dx.doi.org/10.1016/b978-0-08-102894-0.00012-7}
\BIBentrySTDinterwordspacing

\bibitem{Calin2021}
\BIBentryALTinterwordspacing
M.~A. Calin, A.~C. Calin, and D.~N. Nicolae, ``Application of airborne and spaceborne hyperspectral imaging techniques for atmospheric research: past, present, and future,'' \emph{Applied Spectroscopy Reviews}, vol.~56, no.~4, p. 289–323, Apr 2021. [Online]. Available: \url{http://dx.doi.org/10.1080/05704928.2020.1774381}
\BIBentrySTDinterwordspacing

\bibitem{Dale2013}
\BIBentryALTinterwordspacing
L.~M. Dale, A.~Thewis, C.~Boudry, I.~Rotar, P.~Dardenne, V.~Baeten, and J.~A.~F. Pierna, ``\BIBforeignlanguage{en-US}{Hyperspectral imaging applications in agriculture and agro-food product quality and safety control: A review},'' \emph{\BIBforeignlanguage{en-US}{Applied Spectroscopy Reviews}}, vol.~48, no.~2, p. 142–159, Mar 2013. [Online]. Available: \url{http://dx.doi.org/10.1080/05704928.2012.705800}
\BIBentrySTDinterwordspacing

\bibitem{Lu_Dao_Liu_He_Shang_2020}
\BIBentryALTinterwordspacing
B.~Lu, P.~Dao, J.~Liu, Y.~He, and J.~Shang, ``\BIBforeignlanguage{en-US}{Recent advances of hyperspectral imaging technology and applications in agriculture},'' \emph{\BIBforeignlanguage{en-US}{Remote Sensing}}, p. 2659, Aug 2020. [Online]. Available: \url{http://dx.doi.org/10.3390/rs12162659}
\BIBentrySTDinterwordspacing

\bibitem{wang2021review}
C.~Wang, B.~Liu, L.~Liu, Y.~Zhu, J.~Hou, P.~Liu, and X.~Li, ``A review of deep learning used in the hyperspectral image analysis for agriculture,'' \emph{Artificial Intelligence Review}, vol.~54, no.~7, pp. 5205--5253, 2021.

\bibitem{Qin2019Spectral}
Qin, Anyong, Shang, Zhaowei, Tian, Jinyu, Wang, Yulong, Zhang, and Taiping, ``Spectral–spatial graph convolutional networks for semisupervised hyperspectral image classification,'' \emph{IEEE Geoscience and Remote Sensing Letters}, 2019.

\bibitem{2017Deep}
L.~Mou, P.~Ghamisi, and X.~X. Zhu, ``Deep recurrent neural networks for hyperspectral image classification,'' \emph{IEEE Transactions on Geoscience \& Remote Sensing}, pp. 3639--3655, 2017.

\bibitem{Chenyushi2016}
\BIBentryALTinterwordspacing
Y.~Chen, H.~Jiang, C.~Li, X.~Jia, and P.~Ghamisi, ``\BIBforeignlanguage{en-US}{Deep feature extraction and classification of hyperspectral images based on convolutional neural networks},'' \emph{\BIBforeignlanguage{en-US}{IEEE Transactions on Geoscience and Remote Sensing}}, p. 6232–6251, Oct 2016. [Online]. Available: \url{http://dx.doi.org/10.1109/tgrs.2016.2584107}
\BIBentrySTDinterwordspacing

\bibitem{Liy2017}
\BIBentryALTinterwordspacing
Y.~Li, H.~Zhang, and Q.~Shen, ``\BIBforeignlanguage{en-US}{Spectral–spatial classification of hyperspectral imagery with 3d convolutional neural network},'' \emph{\BIBforeignlanguage{en-US}{Remote Sensing}}, p.~67, Jan 2017. [Online]. Available: \url{http://dx.doi.org/10.3390/rs9010067}
\BIBentrySTDinterwordspacing

\bibitem{Zhanghk2019}
\BIBentryALTinterwordspacing
H.~Zhang, Y.~Li, Y.~Jiang, P.~Wang, Q.~Shen, and C.~Shen, ``\BIBforeignlanguage{en-US}{Hyperspectral classification based on lightweight 3-d-cnn with transfer learning},'' \emph{\BIBforeignlanguage{en-US}{IEEE Transactions on Geoscience and Remote Sensing}}, p. 5813–5828, Aug 2019. [Online]. Available: \url{http://dx.doi.org/10.1109/tgrs.2019.2902568}
\BIBentrySTDinterwordspacing

\bibitem{2021An}
A.~Dosovitskiy, L.~Beyer, A.~Kolesnikov, D.~Weissenborn, X.~Zhai, T.~Unterthiner, M.~Dehghani, M.~Minderer, G.~Heigold, and S.~Gelly, ``An image is worth 16x16 words: Transformers for image recognition at scale,'' in \emph{International Conference on Learning Representations}, 2021.

\bibitem{2021SpectralFormer}
D.~Hong, Z.~Han, J.~Yao, L.~Gao, B.~Zhang, A.~Plaza, and J.~Chanussot, ``Spectralformer: Rethinking hyperspectral image classification with transformers,'' 2021.

\bibitem{SunLe}
L.~Sun, G.~Zhao, Y.~Zheng, Z.~Wu, and Y.~Sun, ``\BIBforeignlanguage{en-US}{Spectral-spatial feature tokenization transformer for hyperspectral image classification}.''

\bibitem{HeXin}
X.~He, Y.~Chen, and Q.~Li, ``\BIBforeignlanguage{en-US}{Two-branch pure transformer for hyperspectral image classification}.''

\bibitem{Sun2024}
L.~Sun, X.~Wang, Y.~Zheng, Z.~Wu, and L.~Fu, ``Multiscale 3-d–2-d mixed cnn and lightweight attention-free transformer for hyperspectral and lidar classification,'' \emph{IEEE Transactions on Geoscience and Remote Sensing}, vol.~62, pp. 1--16, 2024.

\bibitem{Saeed2022}
S.~Ghaderizadeh, D.~Abbasi-Moghadam, A.~Sharifi, A.~Tariq, and S.~Qin, ``Multiscale dual-branch residual spectral–spatial network with attention for hyperspectral image classification,'' \emph{IEEE Journal of Selected Topics in Applied Earth Observations and Remote Sensing}, vol.~15, pp. 5455--5467, 2022.

\bibitem{Mei2016}
\BIBentryALTinterwordspacing
S.~Mei, J.~Ji, Q.~Bi, J.~Hou, Q.~Du, and W.~Li, ``\BIBforeignlanguage{en-US}{Integrating spectral and spatial information into deep convolutional neural networks for hyperspectral classification},'' in \emph{\BIBforeignlanguage{en-US}{2016 IEEE International Geoscience and Remote Sensing Symposium (IGARSS)}}, Jul 2016. [Online]. Available: \url{http://dx.doi.org/10.1109/igarss.2016.7730321}
\BIBentrySTDinterwordspacing

\bibitem{Yang2016}
\BIBentryALTinterwordspacing
J.~Yang, Y.~Zhao, J.~C.-W. Chan, and C.~Yi, ``\BIBforeignlanguage{en-US}{Hyperspectral image classification using two-channel deep convolutional neural network},'' in \emph{\BIBforeignlanguage{en-US}{2016 IEEE International Geoscience and Remote Sensing Symposium (IGARSS)}}, Jul 2016. [Online]. Available: \url{http://dx.doi.org/10.1109/igarss.2016.7730324}
\BIBentrySTDinterwordspacing

\bibitem{2017Spectral}
H.~Zhang, Y.~Li, Y.~Zhang, and Q.~Shen, ``Spectral-spatial classification of hyperspectral imagery using a dual-channel convolutional neural network,'' \emph{Remote Sensing Letters}, vol.~8, no. 4-6, pp. 438--447, 2017.

\bibitem{Roy2020}
\BIBentryALTinterwordspacing
S.~K. Roy, G.~Krishna, S.~R. Dubey, and B.~B. Chaudhuri, ``\BIBforeignlanguage{en-US}{Hybridsn: Exploring 3-d–2-d cnn feature hierarchy for hyperspectral image classification},'' \emph{\BIBforeignlanguage{en-US}{IEEE Geoscience and Remote Sensing Letters}}, p. 277–281, Feb 2020. [Online]. Available: \url{http://dx.doi.org/10.1109/lgrs.2019.2918719}
\BIBentrySTDinterwordspacing

\bibitem{Fang2019}
\BIBentryALTinterwordspacing
B.~Fang, Y.~Li, H.~Zhang, and J.~Chan, ``\BIBforeignlanguage{en-US}{Hyperspectral images classification based on dense convolutional networks with spectral-wise attention mechanism},'' \emph{\BIBforeignlanguage{en-US}{Remote Sensing}}, p. 159, Jan 2019. [Online]. Available: \url{http://dx.doi.org/10.3390/rs11020159}
\BIBentrySTDinterwordspacing

\bibitem{Liu2018}
B.~Liu, X.~Yu, P.~Zhang, X.~Tan, R.~Wang, and L.~Zhi, ``Spectral--spatial classification of hyperspectral image using three-dimensional convolution network,'' \emph{Journal of Applied Remote Sensing}, vol.~12, no.~1, pp. 016\,005--016\,005, 2018.

\bibitem{Ladi2022}
\BIBentryALTinterwordspacing
S.~K. Ladi, G.~K. Panda, R.~Dash, and P.~K. Ladi, ``\BIBforeignlanguage{en-US}{A novel strategy for classifying spectral-spatial shallow and deep hyperspectral image features using 1d-ewt and 3d-cnn},'' Aug 2022. [Online]. Available: \url{http://dx.doi.org/10.21203/rs.3.rs-1926705/v1}
\BIBentrySTDinterwordspacing

\bibitem{3dhsi2024}
S.~Y. Chen, K.~H. Hsu, and T.~H. Kuo, ``Hyperspectral target detection-based 2d-3d parallel convolutional neural networks for hyperspectral image classification,'' \emph{IEEE Journal of Selected Topics in Applied Earth Observations and Remote Sensing}, vol.~PP.

\bibitem{Jiang_Li_Zhang_2019}
\BIBentryALTinterwordspacing
Y.~Jiang, Y.~Li, and H.~Zhang, ``\BIBforeignlanguage{en-US}{Hyperspectral image classification based on 3-d separable resnet and transfer learning},'' \emph{\BIBforeignlanguage{en-US}{IEEE Geoscience and Remote Sensing Letters}}, p. 1949–1953, Dec 2019. [Online]. Available: \url{http://dx.doi.org/10.1109/lgrs.2019.2913011}
\BIBentrySTDinterwordspacing

\bibitem{Jiang2023}
M.~Jiang, L.~Gao, A.~Plaza, X.-L. Zhao, X.~Sun, G.~Liu, and Y.~Su, ``\BIBforeignlanguage{en-US}{Graphgst: Graph generative structure-aware transformer for hyperspectral image classification}.''

\bibitem{Graham2021}
\BIBentryALTinterwordspacing
B.~Graham, A.~El-Nouby, H.~Touvron, P.~Stock, A.~Joulin, H.~Jegou, and M.~Douze, ``\BIBforeignlanguage{en-US}{Levit: a vision transformer in convnet’s clothing for faster inference},'' in \emph{\BIBforeignlanguage{en-US}{2021 IEEE/CVF International Conference on Computer Vision (ICCV)}}, Oct 2021. [Online]. Available: \url{http://dx.doi.org/10.1109/iccv48922.2021.01204}
\BIBentrySTDinterwordspacing

\bibitem{Fu2022}
L.~Fu, H.~Tian, X.~Zhai, P.~Gao, and X.~Peng, ``\BIBforeignlanguage{en-US}{Incepformer: Efficient inception transformer with pyramid pooling for semantic segmentation},'' Dec 2022.

\bibitem{Mehta2021}
S.~Mehta and M.~Rastegari, ``\BIBforeignlanguage{en-US}{Mobilevit: Light-weight, general-purpose, and mobile-friendly vision transformer},'' \emph{\BIBforeignlanguage{en-US}{arXiv: Computer Vision and Pattern Recognition,arXiv: Computer Vision and Pattern Recognition}}, Oct 2021.

\bibitem{ZhangH2022}
H.~Zhang, W.~Hu, and X.~Wang, ``\BIBforeignlanguage{en-US}{Parc-net: Position aware circular convolution with merits from convnets and transformer},'' \emph{\BIBforeignlanguage{en-US}{ECCV2022}}, Mar 2022.

\bibitem{Ghaderizadeh2022}
\BIBentryALTinterwordspacing
S.~Ghaderizadeh, D.~Abbasi-Moghadam, A.~Sharifi, A.~Tariq, and S.~Qin, ``\BIBforeignlanguage{en-US}{Multiscale dual-branch residual spectral–spatial network with attention for hyperspectral image classification},'' \emph{\BIBforeignlanguage{en-US}{IEEE Journal of Selected Topics in Applied Earth Observations and Remote Sensing}}, p. 5455–5467, Jan 2022. [Online]. Available: \url{http://dx.doi.org/10.1109/jstars.2022.3188732}
\BIBentrySTDinterwordspacing

\bibitem{He2021}
\BIBentryALTinterwordspacing
X.~He, Y.~Chen, and Z.~Lin, ``\BIBforeignlanguage{en-US}{Spatial-spectral transformer for hyperspectral image classification},'' \emph{\BIBforeignlanguage{en-US}{Remote Sensing}}, p. 498, Jan 2021. [Online]. Available: \url{http://dx.doi.org/10.3390/rs13030498}
\BIBentrySTDinterwordspacing

\bibitem{Chen2021}
Z.~Chen, L.~Xie, J.~Niu, X.~Li, L.~Wei, and Q.~Tian, ``\BIBforeignlanguage{en-US}{Visformer: The vision-friendly transformer},'' \emph{\BIBforeignlanguage{en-US}{Cornell University - arXiv,Cornell University - arXiv}}, Apr 2021.

\bibitem{ZhangFCA2022}
H.~Zhang, W.~Hu, and X.~Wang, ``\BIBforeignlanguage{en-US}{Fcaformer: Forward cross attention in hybrid vision transformer},'' Nov 2022.

\end{thebibliography}
\bibliographystyle{IEEEtran}

\end{document}